\documentclass{paper} 
\usepackage{cite}
\usepackage{amsmath,amssymb,amsfonts}
\usepackage{algorithm}
\usepackage{algorithmic}
\usepackage{graphicx}
\usepackage{textcomp}
\def\BibTeX{{\rm B\kern-.05em{\sc i\kern-.025em b}\kern-.08em
    T\kern-.1667em\lower.7ex\hbox{E}\kern-.125emX}}
    
\begin{document}
\title{Viewpoint-Invariant Exercise Repetition Counting}
\author{Yu Cheng Hsu, Qingpeng Zhang, 
Efstratios Tsougenis, Kwok-Leung Tsui
\thanks{Submit date, fund information go to here }
\thanks{Y.C.H. is with School of Data Science, City University of Hong Kong and A.I Lab, Hospital Authority, Hong Kong }
\thanks{Q.Z. is with School of Data Science, City University of Hong Kong }
\thanks{E.T. is with A.I Lab, Hospital Authority, Hong Kong }
\thanks{K.L.T. is with School of Data Science, City University of Hong Kong and Grado Department of Industrial and Systems Engineering, Virginia Polytechnic Institute and State University, }
}

\maketitle

\begin{abstract}
Counting the repetition of human exercise and physical rehabilitation is a common task in rehabilitation and exercise training. The existing vision-based repetition counting methods less emphasize the concurrent motions in the same video. This work presents a vision-based human motion repetition counting applicable to counting concurrent motions through the skeleton location extracted from various pose estimation methods. The presented method was validated on the University of Idaho Physical Rehabilitation Movements Data Set (UI-PRMD), and MM-fit dataset. The overall mean absolute error (MAE) for mm-fit was 0.06 with off-by-one Accuracy (OBOA) 0.94. Overall MAE for UI-PRMD dataset was 0.06 with OBOA 0.95. We have also tested the performance in different kind of camera location and concurrent motions with conveniently collected video with overall MAE 0.06 and OBOA 0.88. The proposed method provides a view-angle and motion agnostic concurrent motion counting. This method can potentially used in large-scale remote rehabilitation and exercise training with only one camera. 
\end{abstract}

\section{Introduction}
\label{sec:introduction}

Repetitive exercise and training are omnipresent in daily life and, especially, common in sport and rehabilitation training. An early research \cite{jack2010barriers} has surveyed that disobedience of the exercise program is one of the factors that deteriorate the training outcome. Some researchers believed \cite{heath2001increasing,standage2003model} that lack of motivation and feedback is the underlying reason that the users do not comply with the designated exercise plan. Engaging users by providing more feedback and motivating results could increase the adherence of the users, and consequently, enhance the training outcome.   

As the technology advance, training and rehabilitation program might not necessarily take place in the primary care area. The concept of the home exercise program (HEP) and remote rehabilitation with the assistance of information technology draws researchers and medical professionals attention. This approach relief the burden in the medical area, and provide flexibility for people who is difficult in getting medical resources. Nonetheless, the better engagement and adherence in such context is less addressed by the researcher in developing such a solution.

Providing feedback on counting the repetition of the exercise is informative for users. Users can be more engaged in the exercise and training program given the counting feedback, and get a rough idea that they might get fatigued or change the repetition speed during the exercise. Through this feedback, additional incentives and stimuli were provided to the user to encourage the user commits to the program. 

Counting the repetition of exercise is a less studied field comparing with the prosperous growing field in human activity recognition. Dwibedi et al. \cite{dwibedi2020counting} pointed out that it is strenuous to search for suitable video from large-scale public datasets as there is no specific keyword and annotations catering this specific needs. The root cause of this phenomenon is that annotating video/signal across the temporal domain is a labor-intensive and monotonous work. Due to these reasons, the repetition counting task usually encounters the lack of sufficient annotated data. 

The goal of this research is to solve a real-time counting on concurrent human exercise motions. This research provides novel repetition counting methods using skeleton data based on a depth camera. The proposed method first calculate the self-similarity of the sequence of skeleton data. Subsequently, We analyzed the spetrogram of the self-similarity and estimated the count.

The major contribution of this work is that it offers (1) view-angle and motion agnostic vision-based counting methods. This work takes the skeleton data from the motion camera and counting through the temporal similarity. (2) Counting multiple people concurrent repetition in the same frame. The proposed method takes the skeleton data as the input. Current pose estimation methods are capable of estimating multiple people in a frame, so the proposed method can handle multiple people in the same frame and inference the repetition for each person with a different repetition frequency. The current state of art \cite{dwibedi2020counting,Levy2015} usually has the limitation of counting only one motion in the videos. In contrast, through counting by skeleton data, counting on concurrent motion becomes possible.  (3) Real-time inference speed under CPU environment. The proposed model was tested under CPU environment and the inference speed can be at 0.001 seconds per input. Therefore, this method can be adopted in real-time applications. This part of the evaluation was less addressed especially in counting through videos. Nevertheless, real-time feedback for the use for exercise and rehabilitation training can increase the engagement of users. 

\section{Related works}
\label{sec:related_work}
\subsection{Pose estimation}
Vision-based pose estimation is a well-studied field. The goal of pose estimation is to identify the human joint location from the image. Several commercial products and research works are serving this purpose, such as Microsoft Kinect and Martinez et al. \cite{martinez2017simple} for three-dimensional pose estimation. Several developed methods, such as HRnet \cite{SunXLW19,WangSCJDZLMTWLX19,YuanCW19}, OpenPose \cite{Cao2021} and PoseNet \cite{Abadi2016}, also provide pose estimation on the two-dimensional image. These methods detect the head, shoulders, elbows, wrists, waists, knees, and ankles, and indicate the joint location on the image. These methods are able to inference several people in an image with reasonable speed. 

\subsection{Motion repetition counting}
 There are two main approaches in counting motion repetition in terms of data modality. The first category \cite{Soro2019,Shen2018,Muehlbauer2011,Shoaib2016,Morris2014,choi2013automatic,chang2007tracking,mortazavi2014determining} attaches the wearable devices (usually inertial sensors) to the human body and counts the repetition using received signals. Analyzing the temporal repetition of the acceleration signal can be conducted through classical time series analysis. Besides, these wearable devices usually combine with other sensors detecting a variety of vital signs, such as heart rate, electrocardiography (ECG), or blood oxygen, and provide detailed analysis of the exercise. The drawback is that wearable devices need to be attached to each participant and make it expensive for large-scale implementation for only repetition counting purposes. Besides, the wearable device can only count the repetition if the body part where the device is attached involves in the exercise. 

The other approach \cite{Ferreira2020,dwibedi2020counting,Levy2015,Runia2018,stromback2020mm} uses the camera to record the movement and identifies the temporal patterns in the video. Existing research usually solves a more general problem in counting the periodical movements but not necessarily human movements. The core idea of these methods is decomposing it into two sub problems. First, these method will extract features which coincide with the periodicity of the repeated movements. The second step is to developing a counting method base on the extracted features.  These studies either rely on the neural network \cite{Ferreira2020,dwibedi2020counting,Levy2015} or identify the frequency pattern of the feature from computer-vision in the video to extract the features \cite{Runia2018,dwibedi2020counting}. Counting the repetition from the features are done by (1) peak detection, (2) Counting through frequency domain , or (3) neural network methods. Identifying multiple concurrent motions in the video is either less addressed \cite{dwibedi2020counting,Runia2018} or listed as the limitation \cite{Levy2015} in the existing studies. The reason is that these methods leverage this assumption to construct global features in the video. Besides, few do these research works report the inference speed of their proposed method.   

\subsubsection{Extracting temporal periodical features by self-similarity}
 Self-similarity is a common techniques analyzing the periodicity in human motion recognition \cite{junejo2010view,korner2013temporal,sun2015exploring}. These studies have reported that self-similarity exhibited strong robustness to different view angle with appropriate features. Dwibedli et al. \cite{dwibedi2020counting} also pointed out that self-similarity also facilitate an informative information and visual representation for further inspection.   

\subsubsection{Counting on signals}
Counting the self-similarity or any other one-dimensional feature signals can be done through analyzing the frequency domain of the signal\cite{Runia2018,azy2008segmentation,briassouli2007extraction} or peak detection \cite{thangali2005periodic,stromback2020mm}. Peak detection is an intuitive method but heavily depends on the signal quality and fine tuning the hyperparameters in peak detection algorithm. Analyzing the frequency domain is more robust to the quality of signal. Recent research works usually  \cite{Runia2018,azy2008segmentation,briassouli2007extraction} adopt Fourier transform, or wavelet transform to capture the repetitions.   

\subsubsection{Counting through skeleton data}
Ferreira et al. \cite{Ferreira2020} and Stromback et al. \cite{stromback2020mm} studies are similar  to the presented work. It adopted the pose estimation method and extracted features from human joint spatial allocation to classify and count the repetition of the motion. A significant advantage for counting through the skeleton data is that existing pose estimation supports recognizing multiple people in the same frame. Counting through skeleton data enables concurrent motions in the same camera. Nonetheless, this valuable advantage did not explicitly state in the previous study. Besides, Ferreira et al. \cite{Ferreira2020}  works has reported significant differences in the accuracy in different camera locations, and both studies require user to configure different hyperparameters for different kind of exercise.

\section{Method}
\subsection{Data source}
\label{sec:data}

Our method was trained and tested using the UI-PRMD dataset \cite{Vakanski2018} and MM-fit \cite{stromback2020mm}. UI-PRMD dataset \cite{Vakanski2018} is composed of ten motions commonly used in rehabilitation, including (1) deep squat, (2) hurdle step, (3) inline lunge, (4) side lunge, (5) sit to stand, (6) standing active straight leg raise, (7) standing shoulder abduction, (8) standing shoulder extension, (9) standing shoulder internal-external rotation, and (10) standing shoulder scaption. Ten healthy subjects repeated each exercise for 10 times. A Kinect v2 camera was put in front of the subject during the data collection. It is worth to point out that subjects were allowed to use their dominant side performing the tasks, so some might use the right side, and some may use the left side to perform task 2, 3, 4, 6, 7, 8 ,9, and 10.

The MM-fit dataset \cite{stromback2020mm} consist of ten type of commonly used workout for home training
(1) squats, (2) push-ups, (3) shoulder press, (4) lunges, (5) dumbbell rows, (6) sit-ups, (7) triceps extensions, (8) biceps curls, (9) lateral raises, (10) and jumping jacks. There are ten participants in the dataset, each participant was asked to perform a set of exercise which consist several exercise with 10 repetitions. A RGB depth camera was put in front of the participant during exercise. Pose estimation was done by OpenPose for 2D pose estimation, and 3D pose estimation was using the method developed by Martina et al. \cite{martinez2017simple}. 

\subsection{Outline of the proposed method}
\label{sec:model}

The proposed method relied on the existing pose estimation algorithms to extract the skeleton location of each frame. Though this method relied on the pose estimation, but it did not rely on specific type of skeleton format. Consequently, 3D pose skeleton format like  Kinect or 2D skeleton in Openpose are all valid input for this method. After obtaining skeleton data, we computed the pairwise cosine similarity of the skeleton time series. Subsequently, we constructed a spectrogram for the pairwise similarity, and counted the repetition by integrating from the spectrogram.

\subsubsection{Pose Estimation}
 The pose estimation method used in the UI-PRMD dataset \cite{Vakanski2018} was Microsoft Kinect v2 built-in 3D pose estimation, which estimated twenty-five joints locations in the 3D distance from the camera. MM-fit \cite{stromback2020mm} dataset provides 2D pose estimation and 3D pose estimation by Openpose or  Martina et al. \cite{martinez2017simple}. Neither MM-fit nor UI-PRMD datasets provides original video, but it is worth to emphasize that the proposed method is a vision-based method.
 
 \subsubsection{Pairwise cosine similarity calculation}
 
 After getting the skeleton location, we first compute the pairwise cosine similarity of the skeleton data. The reason is that most of the training exercises are back-and-forth movements. We can conceptualize this movement through cosine similarity to quantify it as if it is performing simple harmonic motion. Given a skeleton data time series $X$ with length $t$, $X$ was a matrix with size $(t,j \times d)$, where $j$ was the amount of joint, and $d$ was either 2 or 3 representing the dimension of the data. The self cosine similarity is the normalized dot product between the observation at time $t_0$ and time $t_1$. 

 $$
 \text{self cosine similarity}=\cos(\theta)=\frac{X_{t_0}\cdot X_{t_1}}{\left\Vert X_{t_0} \right\Vert\left\Vert X_{t_1} \right\Vert}
 $$
 
\begin{figure}
\centering{\includegraphics[width=0.48\textwidth]{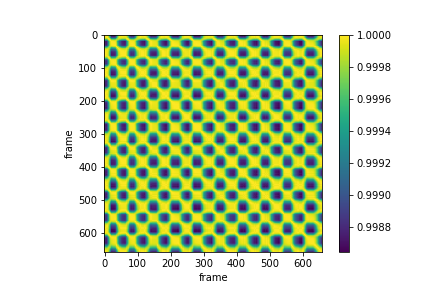}}
\caption{Simulation matrix using one of the clips in MM-fit dataset}
\label{fig:sim-mat}
\end{figure}

 An illustration of output similarity matrix is displayed in Figure \ref{fig:sim-mat}.The mosaic patterns in the Figure \ref{fig:sim-mat} indicated that there is a strong periodicity in the self-similarity.
 
\subsubsection{Constructing spectrogram}
 \begin{figure}
\centering{\includegraphics[width=0.48\textwidth]{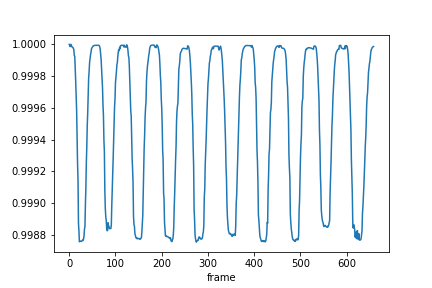}}
\caption{Cosine similarity of frame 0 with rest of the frames}
\label{fig:sim0}
\end{figure}
Observing the strong periodicity, analyzing the frequency patterns and counting through the frequency patterns should be a reasonable approach. We started to analyze the spectrogram of similarity for $t=0$ with respect to rest of the time point $i, \forall i \leq t$. The sequence was demonstrated in Figure \ref{fig:sim0}. The spectrogram was done by calculating the fast Fourier transform (FFT) on a fixed sliding window with window size $w$. For signals $x$ in the sliding window segment, the FFT response: 

$$
\text{FFT}(x)_k=\sum^{w-1}_{n=0}{x(n)\exp(\frac{-i2\pi n}{w})}, k=0,\dots,w-1
$$

\begin{figure}
\centering{\includegraphics[width=0.48\textwidth]{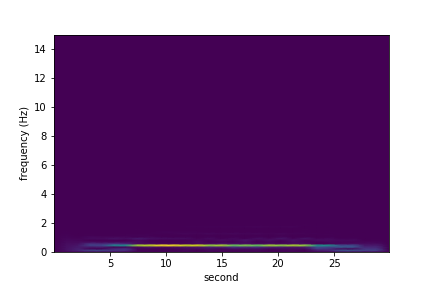}}
\caption{Spectrogram of the cosine similarity sequence}
\label{fig:fft}
\end{figure}

The spectrogram was illustrated in the Figure \ref{fig:fft}. Figure \ref{fig:fft} has revealed that there is a strong signal at the frequency around 0.25 along the whole exercise. 

\subsubsection{Counting out of the time}

We believe the most dominant frequency in the spectrogram is the frequency corresponding to the repeating motion. The local frequency $f_m$ in the sliding window $m$ is the frequency with the highest amplitude in such sliding window. 
$$f_m =\arg\max_k \left\Vert\text{FFT}(x)_k\right\Vert$$

The estimation of the repetition count $\hat{c}$ was defined as the integral of local frequency over time. 

$$
\hat{c}=\sum{t=0}^{T}f_t\delta t
$$

This counting method was adopted in \cite{Runia2018}, which emphasized the capability in counting a non-stationary signals.  
\subsection{Selection of hyperparameter}

The only set of hyperparameter in this method was the sliding window size and sliding steps for spectrogram. There were two factors need to taking into consideration. First, sliding window should be wide enough to envelope at least a cycle of repetition. A cycle of motion usually took around 1 to 3 seconds. Therefore, this will be the lower bound of window size. Besides, the wider the window, the better the frequency resolution. As a trade of, wider window and larger sliding window steps resulted in lower time resolution. We have evaluated and discussed different combination of sliding window size and steps in the result and discussion.

\subsection{Model evaluation}

There are two evaluation metrics commonly used \cite{dwibedi2020counting,Levy2015,Runia2018} in repetition counting.  

\subsubsection{Mean absolute error (MAE) of the count} 
MAE of the count (Equation \ref{eq:maeC}) is the normalized absolute difference between the predicted count $\hat{c}$ and the ground truth $c$. This metric was used in the previous research works as the metric can be interpreted as the percentage of the counting difference comparing with the ground truth. 
\begin{equation}
    \label{eq:maeC}
    \text{MAE of the count} = \frac{1}{n}\sum^n_1{\frac{|\hat{c}-c|}{c}}
\end{equation}

\subsubsection{Off-by-one Accuracy (OBOA)} 
OBOA first labels the video is correctly classified if the absolute difference of predicted count and the ground truth is less or equal to one. Otherwise, it marks the prediction as misclassified. The OBOA is the accuracy using such definition. OBOA provides a brief idea how accurate is the algorithm, and is in favor of accurate algorithm with tolerance of few extremely miscounting cases.

\section{Results and discussion}
\subsection{Motion counting}

\begin{table*}[ht]
    \centering
    \caption{MAE of count and OBOA for motion counting MM-fit dataset, squats (Sq), push ups (Pu), dumbbell shoulder press (Sp), lunges (Lg), dumbbell rows (Dr), situps (Su),tricep extensions (Te),bicep curls (Bc), lateral shoulder raises (Sr), and jumping jacks (Jj) }
    \label{tab:counterror-mmfit}
\begin{tabular}{cccccccccccccc}
   &    &   Sq &   Pu  & Sp &  Lg    & Dr &    Su  & Te & Bc & Sr & Jj & overall MAE & OBOA \\ win & step &            &            &                         &            &               &           &                   &             &            &             &               &           \\\hline
128 & 1  &    0.10 &   0.08 &       0.08 &   0.10 &     0.06 &   0.11 &         0.07 &   0.07 &     0.08 &     0.03 & 0.08  & 0.89 \\    
& 2  &    0.10 &  0.08 & 0.08&   0.10 &      0.06 &   0.11 &    0.07 &   0.07 & 0.08 &     0.03 & 0.08  &  0.89 \\    & 4  &   0.10 &  0.08 &   0.08 &   0.10 &     0.06 &  0.12 &  0.07 &   0.07 & 0.08 &     0.04 & 0.08  &  0.89 \\    & 8  &   0.10 &  0.07 &   0.08 &   0.10 &      0.06 &  0.12 & 0.07 &   0.06 & 0.08 &  0.05& 0.08 &  0.89 \\    
& 16 &   0.10 &  0.07 &     0.06 &  0.10 &      0.07 &  0.12 &  0.07 &   0.07 &     0.08 &  0.06    & 0.08  &  0.89 \\    & 32 &    0.10 &    0.07 &  0.06 &    0.10 & 0.08 &  0.12 &  0.07 &  0.08 &  0.09 &   0.09 & 0.09  &  0.86 \\
256 & 1  &  0.05 &  0.09 &  0.03 &  0.04 &     0.06 &  0.11 &  0.05 &   0.05 &  0.04 &     0.07 & 0.06  &  0.94\\    & 2  &  0.05 &  0.08 & 0.03 &  0.04 &     0.06 &  0.11 &  0.06 &   0.05 &  0.04 &     0.07 & 0.06 &  0.94 \\    & 4  &   0.05 &  0.09 &   0.03 &  0.04 &     0.06 &  0.11 &  0.06 &   0.05 &   0.04 &     0.07 & 0.06 &  0.93 \\    & 8  &  0.05 &  0.09 &    0.04 &  0.04 &     0.06 &  0.11 &  0.06 &   0.05 & 0.04 &     0.07 & 0.06 &  0.92 \\    & 16 &  0.06 &  0.10 &  0.04 &  0.03 &     0.06 &  0.11 & 0.06 &   0.05 &   0.04 &     0.09 & 0.06 &  0.91 \\    & 32 &  0.07 &   0.12 & 0.06 &  0.04 &   0.09 &  0.10 & 0.07 &   0.07 & 0.06 &  0.17 & 0.08 &  0.86 \\
\multicolumn{2}{c}{Baseline \cite{stromback2020mm}} & 0.05 & 0.52 & 0.63 & 0.03 &0.23 & 0.17 &0.33 &4.41 &0.09 &0.26  & 0.67 & 0.90\\

\end{tabular}
\end{table*}
\begin{table*}[ht]
\centering
    \caption{MAE of count and OBOA for motion counting UI-PRMD dataset, deep squat (Dq), hurdle step (Hs), inline lunge (Il), side lunge (Sl), sit to stand (Ss), leg raise (Lr),shoulder abduction (Sa),shoulder extension (Se), shoulder internal-external rotation (Sr), and shoulder scaption (Sc) }
    \label{tab:counterror-UIPRMD}
\begin{tabular}{cccccccccccccc}
    &    & Dq & Hs & Il & Sl & Ss & Lr & Sa & Se & Sr & Sc & overll MAE & OBOA \\
win & step &            &             &              &            &              &           &                    &                    &                            &         &                   &       \\\hline

128 & 1  &       0.07 &         0.10 &          0.10 &       0.14 &         0.12 &      0.15 &               0.08 &               0.11 &                                0.08 &              0.05 & 0.10& 0.81 \\
    & 2  &       0.07 &         0.10 &          0.10 &       0.14 &         0.12 &      0.16 &               0.08 &               0.11 &                                0.08 &              0.05 &0.10&   0.8 \\
    & 4  &       0.08 &        0.09 &          0.10 &       0.14 &         0.11 &      0.15 &               0.09 &               0.11 &                                0.09 &              0.04 &0.10&  0.82 \\
    & 8  &       0.09 &        0.09 &          0.10 &       0.12 &         0.11 &      0.15 &               0.09 &               0.11 &                                0.09 &              0.04 &0.10&  0.82 \\
    & 16 &       0.08 &        0.07 &         0.09 &       0.12 &         0.09 &      0.14 &               0.09 &               0.12 &                                0.09 &              0.03 &0.09&  0.87 \\
    & 32 &       0.07 &        0.07 &         0.09 &       0.13 &         0.09 &      0.13 &               0.07 &                0.10 &                                0.08 &              0.03 &0.08&  0.86 \\
256 & 1  &       0.05 &           0.00 &         0.04 &       0.07 &         0.02 &      0.11 &               0.05 &               0.12 &                                0.08 &              0.06 &0.06&  0.94 \\
    & 2  &       0.05 &        0.01 &         0.04 &       0.07 &         0.02 &      0.11 &               0.05 &               0.12 &                                0.08 &              0.06 &0.06&  0.94 \\
    & 4  &       0.05 &        0.01 &         0.04 &       0.07 &         0.02 &       0.10 &               0.05 &               0.12 &                                0.08 &              0.06&0.06 &  0.94 \\
    & 8  &       0.05 &        0.01 &         0.04 &       0.07 &         0.04 &       0.10 &               0.06 &               0.12 &                                0.07 &              0.05 &0.06&  0.94 \\
    & 16 &       0.04 &        0.02 &         0.02 &       0.05 &         0.05 &      0.11 &               0.05 &               0.13 &                                0.05 &              0.03 &0.06&  0.94 \\
    & 32 &       0.04 &        0.06 &         0.05 &       0.04 &         0.05 &       0.10 &               0.03 &               0.14 &                                0.05 &              0.05 &0.06&  0.95 \\
\end{tabular}
\end{table*}
Overall, the proposed method can reach 0.94 to 0.95 OBOA with MAE 0.06 with the best hyperparameter combinations.
\subsubsection{MM-fit dataset}
Table \ref{tab:counterror-mmfit} indicate the MAE and OBOA for the proposed method and the baseline demonstrated in \cite{stromback2020mm}. The MAE and OBO error was reported with the same testing scheme as Stromback et al. \cite{stromback2020mm}. The proposed method yielded same or significantly better in each individual task. In addition to the baseline method, the proposed method performs more stably across different kind of motion. In contrast, the baseline method vary significantly among different kind of exercise. This might due to the baseline method needs to adjust the hyperparameter specific to the motion.

\subsubsection{UI-PRMD dataset}
Table \ref{tab:counterror-UIPRMD} indicate the MAE and OBOA for the proposed method testing on UI-PRMD dataset. All the data were used to report the result. To the best of our knowledge, there is no research work adopted this dataset for motion counting.

\subsection{Effect of hyperparameters}

Table \ref{tab:counterror-mmfit} and \ref{tab:counterror-UIPRMD} also demonstrated the effect of hyperparameters. We could observe that window size of 256 is more desirable than 128. window size with length 256 perform better than window size with 128 in overall MAE and OBOA. We suspected the reason might be that 256 window size offer higher frequency resolution than 128 window. This difference will be accumulated and amplified along the integration calculation and deteriorate counting accuracy.  

Sliding window step size slightly affected the repetition counting result. Table \ref{tab:counterror-mmfit} showed that MAE increased and OBOA decreased while increasing the step size in MM-fit dataset. Interestingly, this effect cannot be observed in the UI-PRMD dataset (Table \ref{tab:counterror-UIPRMD}. In any case, the result shown in Table \ref{tab:counterror-mmfit} and \ref{tab:counterror-UIPRMD}  indicated that sliding window steps only made minor difference in repetition counting.

\subsection{Counting in the real word}

We acknowledged that both public dataset fixed the camera location and did not tested with concurrent motions in the same videos. Five adults were recruited conveniently and asked to squat for ten repetitions with different camera angle or concurrently. The detailed decomposition for experiment is listed in Table \ref{tab:exp}. Participants one to three are asked to squat for ten times for each camera location. Participants 4 and 5 are asked to squat together with camera in the front. Participant 4 was facing to the camera, and the participant 5 was facing to the right. Videos were recorded by iPad Pro 5th generation, and the pose estimation was made by OpenPose \cite{Cao2021}.

\begin{table}[htp]
    \centering
    \caption{Inference time for each motion}
    \label{tab:exp}
    \begin{tabular}{c c }
 Participants ID & camera location  \\\hline
    1   &  left, front, right \\
  2   &  left, front, right \\
   3   &  left, front, right \\
    4,5  & front,right\\ 
    \end{tabular}
\end{table}

The result in Table \ref{tab:real} was obtained using 256 window size with step 1. Table \ref{tab:real} indicated that the proposed method are able to count with different view angle and concurrent motions. Squats conducted by participant 3, 4 and 5 are correctly counted regadless of view angle or concurrent motions in the Video 10. Minor miscounting happened only in participant 1 and 2.  

\begin{table}[htp]
    \centering
    \caption{Inference time for each motion}
    \label{tab:real}
    \begin{tabular}{c c c c}
 Video ID & Participant ID & Camera location & MAE    \\\hline
    1   &  1 & front & 0.10\\
  2   &  1 & right & 0.10\\
  3   &  1 & left & 0.00\\
    4   &  2 & front & 0.20\\
  5   &  2 & right & 0.00\\
  6   &  2 & left & 0.10\\
    7   &  3 & front & 0.00\\
 8   &  3 & right & 0.00\\
  9   &  3 & left & 0.00\\
  10   &  4 & front & 0.00\\
  10   &  5 & right & 0.00\\
  \hline
  \multicolumn{3}{c}{Overall MAE} & 0.06\\
\multicolumn{3}{c}{OBOA} & 0.88\\
    \end{tabular}
\end{table}

\subsection{Counting in different type of skeleton formats}
\begin{table*}[h]
    \centering
    \caption{MAE of count and OBOA for motion counting MM-fit dataset (2D), squats (Sq), push ups (Pu), dumbbell shoulder press (Sp), lunges (Lg), dumbbell rows (Dr), situps (Su),tricep extensions (Te),bicep curls (Bc), lateral shoulder raises (Sr), and jumping jacks (Jj) }
    \begin{tabular}{cccccccccccccc}
         &    &   Sq &   Pu  & Sp &  Lg    & Dr &    Su  & Te & Bc & Sr & Jj & overall MAE & OBOA \\
    win & step &         &          &                          &         &                &         &                    &              &                          &                &       \\
    128 & 1  &    0.11 &     0.10 &                     0.10 &    0.10 &           0.07 &    0.22 &               0.07 &         0.07 &                     0.08 &           0.04 &0.09&  0.87 \\
        & 2  &    0.11 &     0.10 &                     0.10 &    0.10 &           0.07 &    0.22 &               0.07 &         0.07 &                     0.08 &           0.04 &0.09&  0.87 \\
        & 4  &    0.10 &     0.10 &                     0.09 &    0.10 &           0.06 &    0.21 &               0.08 &         0.07 &                     0.08 &           0.04 &0.09&  0.88 \\
        & 8  &    0.11 &     0.09 &                     0.09 &    0.10 &           0.06 &    0.21 &               0.08 &         0.07 &                     0.08 &           0.05 &0.09&  0.88 \\
        & 16 &    0.11 &     0.09 &                     0.07 &    0.10 &           0.07 &    0.21 &               0.08 &         0.07 &                     0.08 &           0.06 &0.09&  0.87 \\
        & 32 &    0.11 &     0.09 &                     0.08 &    0.10 &           0.07 &    0.20 &               0.08 &         0.08 &                     0.09 &           0.09 &0.10&  0.85 \\
    256 & 1  &    0.06 &     0.13 &                     0.04 &    0.04 &           0.06 &    0.17 &               0.06 &         0.05 &                     0.04 &           0.13 &0.08&  0.87 \\
        & 2  &    0.06 &     0.14 &                     0.04 &    0.04 &           0.06 &    0.17 &               0.06 &         0.05 &                     0.04 &           0.14 &0.08&  0.87 \\
        & 4  &    0.06 &     0.13 &                     0.04 &    0.04 &           0.05 &    0.17 &               0.06 &         0.05 &                     0.04 &           0.14 &0.08&  0.87 \\
        & 8  &    0.05 &     0.14 &                     0.05 &    0.04 &           0.04 &    0.16 &               0.06 &         0.05 &                     0.04 &           0.14 &0.08&  0.86 \\
        & 16 &    0.07 &     0.14 &                     0.05 &    0.04 &           0.05 &    0.17 &               0.06 &         0.05 &                     0.03 &           0.15 &0.08&  0.85 \\
        & 32 &    0.08 &     0.18 &                     0.08 &    0.05 &           0.09 &    0.17 &               0.07 &         0.07 &                     0.06 &           0.20 &0.10&  0.81 \\
    \end{tabular}

    \label{tab:counterror-mmfit2d}
\end{table*}

We have also illustrated counting in different type of skeleton format. MM-fit data is either 3D COCO skeleton format (18 joints, 3D) or 2D COCO skeleton format (18 joints, 2D). UI-PRMD dataset is using Microsoft Kinect skeleton format which has 22 3D joints location. The data collected in this study is in 2D COCO skeleton format. The proposed method all exhibited satisfactory result. We have listed the performance of MM-fit dataset using 2D pose estimation format as the input in Table \ref{tab:counterror-mmfit2d}. Table \ref{tab:counterror-mmfit2d} indicated that counting based on the 2D skeleton format will deteriorate the OBOA and MAE comparing with 3D pose estimation inputs.
 We could observe that counting in the 2D  will deteriorate the result comparing with using 3D results. This deterioration is especially obvious in push ups, situps and jumping jacks. The reason for this variance among different exercise is not that clear, and firther investigation might be helpful.  

\subsection{Limitation and future research direction }
This method assumes the motion is a simple back and forth exercise which the similarity of skeleton data behave like a periodical signals. Exercise with more complicated procedure might not demonstrate this property. Nonetheless, most of the rehab or exercise training does not consist this kind of complex exercise. Second, we assume the most dominant frequency is the frequency corresponding to the exercise. Minor motion counting will need further investigation on the spectrogram. The last limitation is that we rely on motion detection and recognition repetition motion to perform such algorithm. We might investigate more kind of exercise in different context to understand how these limitation affect the proposed method.

\section{Conclusion}
We have presented a motion repetition counting method using skeleton data. This research provide a solution to concurrent repetition counting in human exercise, which is less addressed in the previous study. The proposed method has been verified on the public datasets, and few conveniently collected videos with desirable accuracy in different view angle, motion, and concurrent motions. The proposed method possesses the potential in rehabilitation and exercise training to monitor the progress of the exercise training and provides remote rehabilitation and exercise training. 

\bibliography{generic-color-brief}

\begin{thebibliography}{10}
\providecommand{\url}[1]{#1}
\csname url@samestyle\endcsname
\providecommand{\newblock}{\relax}
\providecommand{\bibinfo}[2]{#2}
\providecommand{\BIBentrySTDinterwordspacing}{\spaceskip=0pt\relax}
\providecommand{\BIBentryALTinterwordstretchfactor}{4}
\providecommand{\BIBentryALTinterwordspacing}{\spaceskip=\fontdimen2\font plus
\BIBentryALTinterwordstretchfactor\fontdimen3\font minus
  \fontdimen4\font\relax}
\providecommand{\BIBforeignlanguage}[2]{{%
\expandafter\ifx\csname l@#1\endcsname\relax
\typeout{** WARNING: IEEEtran.bst: No hyphenation pattern has been}%
\typeout{** loaded for the language `#1'. Using the pattern for}%
\typeout{** the default language instead.}%
\else
\language=\csname l@#1\endcsname
\fi
#2}}
\providecommand{\BIBdecl}{\relax}
\BIBdecl

\bibitem{jack2010barriers}
K.~Jack, S.~M. McLean, J.~K. Moffett, and E.~Gardiner, ``Barriers to treatment
  adherence in physiotherapy outpatient clinics: a systematic review,''
  \emph{Manual therapy}, vol.~15, no.~3, pp. 220--228, 2010.

\bibitem{heath2001increasing}
G.~Heath, E.~H. Howze, E.~B. Kahn, and L.~T. Ramsey, ``Increasing physical
  activity; a report on recommendations of the task force on community
  preventive services,'' 2001.

\bibitem{standage2003model}
M.~Standage, J.~L. Duda, and N.~Ntoumanis, ``A model of contextual motivation
  in physical education: Using constructs from self-determination and
  achievement goal theories to predict physical activity intentions.''
  \emph{Journal of educational psychology}, vol.~95, no.~1, p.~97, 2003.

\bibitem{dwibedi2020counting}
D.~Dwibedi, Y.~Aytar, J.~Tompson, P.~Sermanet, and A.~Zisserman, ``Counting out
  time: Class agnostic video repetition counting in the wild,'' in
  \emph{Proceedings of the IEEE/CVF Conference on Computer Vision and Pattern
  Recognition}, 2020, pp. 10\,387--10\,396.

\bibitem{Levy2015}
O.~Levy and L.~Wolf, ``Live repetition counting,'' 2015.

\bibitem{martinez2017simple}
J.~Martinez, R.~Hossain, J.~Romero, and J.~J. Little, ``A simple yet effective
  baseline for 3d human pose estimation,'' in \emph{Proceedings of the IEEE
  International Conference on Computer Vision}, 2017, pp. 2640--2649.

\bibitem{SunXLW19}
K.~Sun, B.~Xiao, D.~Liu, and J.~Wang, ``Deep high-resolution representation
  learning for human pose estimation,'' in \emph{CVPR}, 2019.

\bibitem{WangSCJDZLMTWLX19}
J.~Wang, K.~Sun, T.~Cheng, B.~Jiang, C.~Deng, Y.~Zhao, D.~Liu, Y.~Mu, M.~Tan,
  X.~Wang, W.~Liu, and B.~Xiao, ``Deep high-resolution representation learning
  for visual recognition,'' \emph{TPAMI}, 2019.

\bibitem{YuanCW19}
Y.~Yuan, X.~Chen, and J.~Wang, ``Object-contextual representations for semantic
  segmentation,'' 2020.

\bibitem{Cao2021}
Z.~Cao, G.~Hidalgo, T.~Simon, S.~E. Wei, and Y.~Sheikh, ``Openpose: Realtime
  multi-person 2d pose estimation using part affinity fields,'' \emph{IEEE
  Transactions on Pattern Analysis and Machine Intelligence}, 2021.

\bibitem{Abadi2016}
M.~Abadi, P.~Barham, J.~Chen, Z.~Chen, A.~Davis, J.~Dean, M.~Devin,
  S.~Ghemawat, G.~Irving, M.~Isard, M.~Kudlur, J.~Levenberg, R.~Monga,
  S.~Moore, D.~G. Murray, B.~Steiner, P.~Tucker, V.~Vasudevan, P.~Warden,
  M.~Wicke, Y.~Yu, and X.~Zheng, ``Tensorflow: A system for large-scale machine
  learning,'' 2016.

\bibitem{Soro2019}
A.~Soro, G.~Brunner, S.~Tanner, and R.~Wattenhofer, ``Recognition and
  repetition counting for complex physical exercises with deep learning,''
  \emph{Sensors (Switzerland)}, 2019.

\bibitem{Shen2018}
C.~Shen, B.~J. Ho, and M.~Srivastava, ``Milift: Efficient smartwatch-based
  workout tracking using automatic segmentation,'' \emph{IEEE Transactions on
  Mobile Computing}, 2018.

\bibitem{Muehlbauer2011}
M.~Muehlbauer, G.~Bahle, and P.~Lukowicz, ``What can an arm holster worn smart
  phone do for activity recognition?'' 2011.

\bibitem{Shoaib2016}
M.~Shoaib, S.~Bosch, O.~D. Incel, H.~Scholten, and P.~J. Havinga, ``Complex
  human activity recognition using smartphone and wrist-worn motion sensors,''
  \emph{Sensors (Switzerland)}, 2016.

\bibitem{Morris2014}
D.~Morris, T.~S. Saponas, A.~Guillory, and I.~Kelner, ``Recofit: Using a
  wearable sensor to find, recognize, and count repetitive exercises,'' 2014.

\bibitem{choi2013automatic}
K.~S. Choi, Y.~S. Joo, and S.-K. Kim, ``Automatic exercise counter for outdoor
  exercise equipment,'' in \emph{2013 IEEE International Conference on Consumer
  Electronics (ICCE)}.\hskip 1em plus 0.5em minus 0.4em\relax IEEE, 2013, pp.
  436--437.

\bibitem{chang2007tracking}
K.-h. Chang, M.~Y. Chen, and J.~Canny, ``Tracking free-weight exercises,'' in
  \emph{International Conference on Ubiquitous Computing}.\hskip 1em plus 0.5em
  minus 0.4em\relax Springer, 2007, pp. 19--37.

\bibitem{mortazavi2014determining}
B.~J. Mortazavi, M.~Pourhomayoun, G.~Alsheikh, N.~Alshurafa, S.~I. Lee, and
  M.~Sarrafzadeh, ``Determining the single best axis for exercise repetition
  recognition and counting on smartwatches,'' in \emph{2014 11th International
  Conference on Wearable and Implantable Body Sensor Networks}.\hskip 1em plus
  0.5em minus 0.4em\relax IEEE, 2014, pp. 33--38.

\bibitem{Ferreira2020}
B.~Ferreira, P.~M. Ferreira, G.~Pinheiro, N.~Figueiredo, F.~Carvalho,
  P.~Menezes, and J.~Batista, ``Exploring workout repetition counting and
  validation through deep learning,'' 2020.

\bibitem{Runia2018}
T.~F.~H. Runia, C.~G.~M. Snoek, and A.~W.~M. Smeulders, ``Real-world repetition
  estimation by div, grad and curl,'' \emph{cvpr}, 2018.

\bibitem{stromback2020mm}
D.~Str{\"o}mb{\"a}ck, S.~Huang, and V.~Radu, ``Mm-fit: Multimodal deep learning
  for automatic exercise logging across sensing devices,'' \emph{Proceedings of
  the ACM on Interactive, Mobile, Wearable and Ubiquitous Technologies},
  vol.~4, no.~4, pp. 1--22, 2020.

\bibitem{junejo2010view}
I.~N. Junejo, E.~Dexter, I.~Laptev, and P.~Perez, ``View-independent action
  recognition from temporal self-similarities,'' \emph{IEEE transactions on
  pattern analysis and machine intelligence}, vol.~33, no.~1, pp. 172--185,
  2010.

\bibitem{korner2013temporal}
M.~K{\"o}rner and J.~Denzler, ``Temporal self-similarity for appearance-based
  action recognition in multi-view setups,'' in \emph{International Conference
  on Computer Analysis of Images and Patterns}.\hskip 1em plus 0.5em minus
  0.4em\relax Springer, 2013, pp. 163--171.

\bibitem{sun2015exploring}
C.~Sun, I.~N. Junejo, M.~Tappen, and H.~Foroosh, ``Exploring sparseness and
  self-similarity for action recognition,'' \emph{IEEE Transactions on Image
  Processing}, vol.~24, no.~8, pp. 2488--2501, 2015.

\bibitem{azy2008segmentation}
O.~Azy and N.~Ahuja, ``Segmentation of periodically moving objects,'' in
  \emph{2008 19th International Conference on Pattern Recognition}.\hskip 1em
  plus 0.5em minus 0.4em\relax IEEE, 2008, pp. 1--4.

\bibitem{briassouli2007extraction}
A.~Briassouli and N.~Ahuja, ``Extraction and analysis of multiple periodic
  motions in video sequences,'' \emph{IEEE transactions on pattern analysis and
  machine intelligence}, vol.~29, no.~7, pp. 1244--1261, 2007.

\bibitem{thangali2005periodic}
A.~Thangali and S.~Sclaroff, ``Periodic motion detection and estimation via
  space-time sampling,'' in \emph{2005 Seventh IEEE Workshops on Applications
  of Computer Vision (WACV/MOTION'05)-Volume 1}, vol.~2.\hskip 1em plus 0.5em
  minus 0.4em\relax IEEE, 2005, pp. 176--182.

\bibitem{Vakanski2018}
A.~Vakanski, H.~P. Jun, D.~Paul, and R.~Baker, ``A data set of human body
  movements for physical rehabilitation exercises,'' \emph{Data}, 2018.

\end{thebibliography}
\bibliographystyle{IEEEtran}
\end{document}